\documentclass{article}

\usepackage{microtype}
\usepackage{graphicx}
\usepackage{subcaption}
\usepackage{booktabs}
\usepackage{hyperref}
\usepackage{url}
\usepackage{amsfonts}
\usepackage{amsmath}
\usepackage{amssymb}
\usepackage{nicefrac}
\usepackage{xcolor}
\usepackage{algorithm}
\usepackage{algorithmic}
\usepackage{multirow}
\usepackage{colortbl}
\usepackage{enumitem}
\usepackage{tikz}
\usetikzlibrary{shapes.geometric,arrows.meta,positioning,fit,backgrounds,calc}
\usepackage{balance}

\usepackage[accepted]{icml2026}

\newcommand{\ours}{\textsc{NKI-Agent}}
\newcommand{\nkibench}{\textsc{NKIGen-Bench}}
\newcommand{\nki}{\textsc{NKI}}

\icmltitlerunning{NKI-Agent: Agentic Tool Use for Neuron Kernel Generation}

\begin{document}

\twocolumn[
\icmltitle{NKI-Agent: Domain-Specific Fine-Tuning and Agentic Tool Use\\
for Neuron Kernel Generation}

\begin{icmlauthorlist}
\icmlauthor{Junjie Tang}{aws}
\icmlauthor{Jun Huan}{aws}
\icmlauthor{Hao Zhou}{aws}
\icmlauthor{Yuhao Zhang}{aws}
\icmlauthor{Lin Wang}{aws}
\end{icmlauthorlist}

\icmlaffiliation{aws}{Amazon Web Services}

\icmlcorrespondingauthor{Junjie Tang}{junjtang@amazon.com}

\icmlkeywords{coding agents, tool use, kernel generation, AWS Trainium,
              Neuron Kernel Interface, supervised fine-tuning, GRPO,
              agentic workflows, code LLMs}

\vskip 0.3in
]

\printAffiliationsAndNotice{}

\begin{abstract}
Recent agentic approaches to LLM-based kernel generation have
achieved impressive results on CUDA. For emerging AI accelerators
such as AWS Trainium and Inferentia, automated kernel generation and optimization remain largely unaddressed. Writing
kernels for these chips via the Neuron Kernel Interface (\nki{})
is particularly challenging: developers must navigate a multi-engine architecture, tile-based programming, and explicit data movement across multi-level memory hierarchy. Moreover, no publicly-available training
data, benchmarks, or tool-augmented agents exist for this domain.
We introduce \ours{}, the first system combining domain-specific
supervised fine-tuning (SFT) with a compile-verify-fix agent
loop for \nki{} kernel generation. We adapt the existing
CUDA-Agent framework to Neuron hardware, curate 6{,}000 \nki{}
kernel generation tasks for training, and construct
\nkibench{}, a 250-task benchmark across three difficulty
levels. Evaluated on real Trn1 hardware, \ours{} with Claude Opus~4.8
and a rank-aware system prompt achieves a \textbf{77.3\% pass
rate} on the 150-task \nkibench{}. We show that tools use is critical: Opus~4.8 scores 6\% in single-shot mode without agent tools. On a 60-task subset, we show that an SFT-trained Qwen3-Coder-30B-A3B achieves 25.0\% pass
rate at \textbf{1/100th the cost}, outperforming
Claude Sonnet~4 (15.0\%). We also report that Group Relative Policy Optimization
(GRPO) with binary compilation reward
fails to improve over SFT, providing guidance on reward design
for RL-based kernel generation.
\end{abstract}

\section{Introduction}
\label{sec:introduction}

The rapid scaling of deep learning models has intensified demand
for hardware-specific kernel optimization. While NVIDIA GPUs have
benefited from decades of CUDA ecosystem development and compiler
optimizations, emerging AI accelerators such as AWS Trainium and
Inferentia require new approaches to kernel generation. The
Neuron Kernel Interface (\nki{})~\cite{nki_docs} provides a
Python-based programming model for these chips, but writing
high-performance \nki{} kernels remains challenging due to the
unique multi-engine architecture and tile-based programming model.

Recent work on agentic code generation has demonstrated that
agents equipped with execution feedback can
generate competitive GPU kernels. The CUDA-Agent~\cite{cuda_agent}
system uses Proximal Policy Optimization (PPO)-trained agents on a synthesized dataset of
6{,}000 CUDA tasks, achieving 92--100\% faster rates over
\texttt{torch.compile} on KernelBench~\cite{kernelbench}. However,
this approach is specific to CUDA and does not address the
distinct challenges of Neuron hardware.

\nki{} programming presents unique challenges:
(1)~\textbf{Multi-engine architecture} with four specialized
compute engines (Tensor, Vector, Scalar, GpSimd);
(2)~\textbf{Tile-based programming} on 2D tiles with fixed
128-element partition dimension;
(3)~\textbf{Explicit memory hierarchy} requiring manual data
movement between HBM (high-bandwidth memory), SBUF
(on-chip scratchpad), and PSUM (partial-sum accumulator);
(4)~\textbf{Evolving SDK} with API changes between versions.

In this paper, we present \ours{}, the first system to apply
domain-specific fine-tuning and agentic tool use to \nki{}
kernel generation. Our contributions:

\begin{enumerate}[leftmargin=*,itemsep=2pt]
    \item \textbf{\ours{}}: A multi-turn agent with compile and
    verify tools and a rank-aware system prompt for \nki{}
    kernel generation. It supports various models including a fine-tuned Qwen3-Coder-30B-A3B model with supervised fine-tuning
    (SFT) and Group Relative Policy Optimization (GRPO).
    \item \textbf{NKI-Agent-Ops-6K \& \nkibench{}}: 6{,}000
    curated \nki{} tasks and a 250-task benchmark across three
    difficulty levels.
    \item \textbf{SFT data quality analysis}: Four dataset
    iterations showing data curation matters more than
    hyperparameters.
    \item \textbf{Negative result}: GRPO with binary
    reward fails to improve over SFT.
\end{enumerate}

The \ours{} framework achieves 77.3\% on the 150-task
\nkibench{} with a frontier model (Opus~4.8 with a rank-aware
prompt), and our domain-specialized SFT model captures a large
fraction of this performance at 1/100th the cost (25.0\% vs
63.3\% on the 60-task subset), outperforming Sonnet~4 (15.0\%)
with identical tools.

\section{Related work}
\label{sec:related}

\paragraph{LLM-based kernel generation.}
KernelBench~\cite{kernelbench} benchmarks LLM-generated GPU
kernels. CUDA-Agent~\cite{cuda_agent} trains PPO-based agents on
6{,}000 CUDA tasks with compile-verify-profile tools. Other
approaches include AlphaCode-style search~\cite{li2022alphacode},
self-repair~\cite{olausson2024selfrepair}, and
TritonBench~\cite{tritonbench}. Related code-translation work
generates accelerator code by translating across languages:
TransCoder~\cite{roziere2020transcoder},
BabelTower~\cite{wen2022babeltower}, and
CodeRosetta~\cite{tehrani2024coderosetta}. The agentic-coding
evaluation lineage (SWE-bench~\cite{jimenez2023swebench},
SWE-agent~\cite{yang2024sweagent}) motivates our
tool-augmented, execution-grounded setup.

\paragraph{RL for code generation.}
CodeRL~\cite{le2022coderl} and PPOCoder~\cite{shojaee2023ppocoder}
use execution rewards; DeepSeek-R1~\cite{deepseekr1} demonstrates
GRPO~\cite{shao2024deepseek} without a critic. CUDA-Agent shows RL
with graded reward outperforms SFT; we show the converse for
binary reward.

\paragraph{Neuron hardware and NKI.}
AWS Trainium/Inferentia chips~\cite{aws_neuron} feature four
specialized compute engines. \nki{}~\cite{nki_docs} provides
tile-based programming; prior work is limited to reference
kernels~\cite{nki_samples}. AccelOpt~\cite{accelopt}
(concurrent) addresses \nki{} kernel \emph{optimization},
complementary to our correctness-first generation.

\section{Method}
\label{sec:method}

\begin{figure*}[t]
    \centering
    \includegraphics[width=\textwidth]{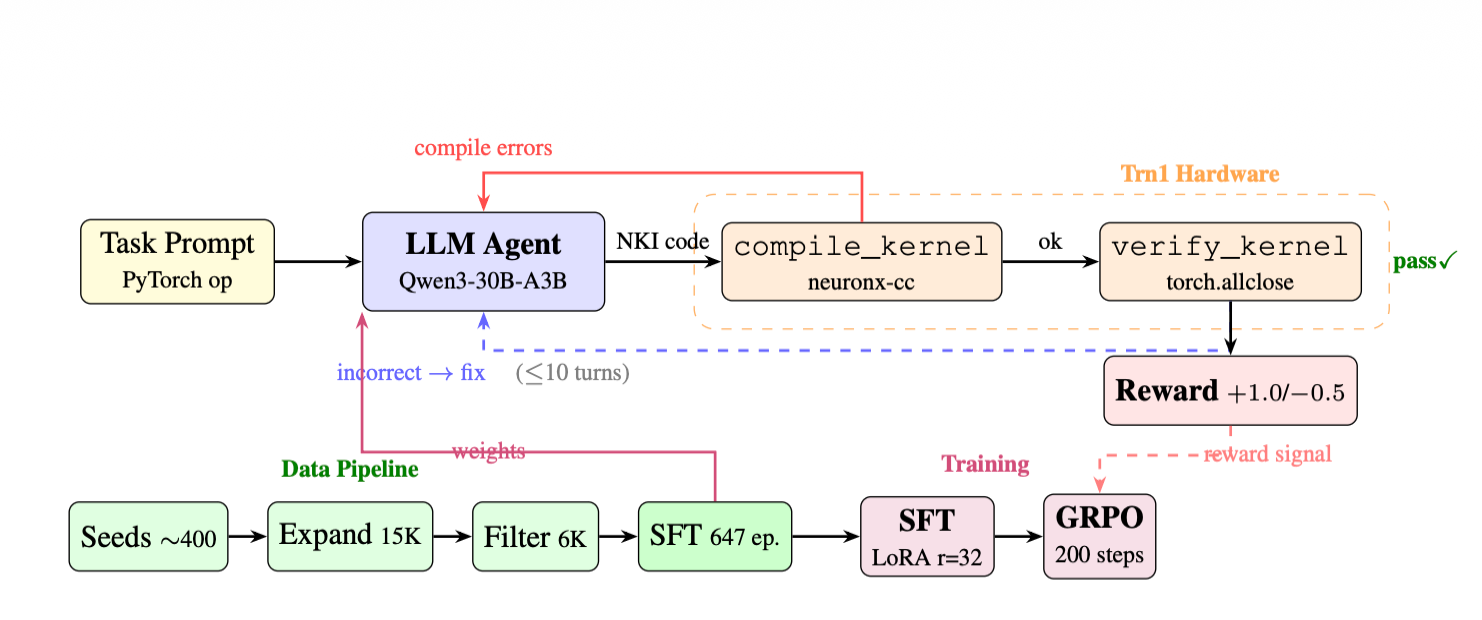}
    \caption{\textbf{\ours{} system overview.} \emph{Top:} The
    agent generates \nki{} code, then iteratively compiles and
    verifies on Trn1 hardware ($\leq$10 turns). Compile errors
    and verification failures feed back for correction.
    \emph{Bottom:} Data pipeline produces 6{,}000 tasks; SFT on
    curated episodes followed by GRPO with binary reward.}
    \label{fig:architecture}
\end{figure*}

Figure~\ref{fig:architecture} illustrates the \ours{} system. It
consists of four components: (1)~a data synthesis pipeline
producing 6{,}000 curated \nki{} tasks, (2)~a multi-turn agent
with compile and verify tools, (3)~a binary reward function, and
(4)~a two-stage SFT $\rightarrow$ GRPO training pipeline.

\subsection{NKI programming model}
\label{sec:nki_model}

Neuron chips feature four specialized engines: \textbf{Tensor}
(matrix ops), \textbf{Vector} (elementwise/reduction),
\textbf{Scalar} (control flow), and \textbf{GpSimd}
(general-purpose). Programs operate on 2D tiles with a fixed
128-element partition dimension. Data must be explicitly moved
between HBM (device memory), SBUF (24~MB scratchpad), and PSUM
(partial sum accumulator) in a load, compute, store
pattern. This differs fundamentally from CUDA's thread-block
model, requiring domain-specific knowledge for correct kernels.

\subsection{Data synthesis pipeline}
\label{sec:data_pipeline}

We synthesize a dataset NKI-Agent-Ops-6K in three stages:
(1)~\textbf{Seed crawling} from PyTorch \texttt{torch.nn}
($\sim$180 ops), \texttt{nki-samples} ($\sim$30 kernels), and
KernelBench ($\sim$250 tasks), yielding $\sim$400 seeds.
(2)~\textbf{Combinatorial expansion} via shape ($5\times$),
dtype ($3\times$), and fusion synthesis ($\sim$200 patterns),
producing $\sim$15K candidates.
(3)~\textbf{Filtering} through compilation, execution, stability
(CV$<$10\%), and baseline checks, balanced to 6{,}000 tasks. The
250 \nkibench{} tasks are held out. For SFT, we curate 647
high-quality episodes through four iterations
(Section~\ref{sec:sft_data_quality}).

\subsection{Agent architecture}
\label{sec:agent}

\ours{} operates through multi-turn interaction with two tools
on real Trn1 hardware: (1)~\textbf{compile\_kernel} invokes
\texttt{neuronx-cc} with 120s timeout and returns compilation status with error
messages; (2)~\textbf{verify\_kernel} runs the
compiled kernel with random inputs, comparing against the
PyTorch reference via \texttt{torch.allclose}. The agent iterates up to
$T{=}10$ turns: generate, compile, verify, fix.
We set $T{=}10$ empirically: in our evaluation runs the vast
majority of eventual successes converge within the first few
turns, and gains beyond ten turns are negligible while the
per-turn Trn1 compile/verify cost grows linearly, so $T{=}10$
balances solution coverage against evaluation cost.

\paragraph{Rank-aware system prompt.}
NKIBench inputs span 2-D, 3-D, and 4-D tensors (e.g.\ batched
attention and N,C,H,W convolutions), but a naive prompt assumes
a 2-D $(M, N)$ layout. We use a rank-aware prompt that instructs the model
to read each task's input-shape block and provides templates for
2-D, 3-D, and 4-D kernels, plus a reminder that every output
element must be written by an \texttt{nl.store}. This is a
prompt-only change (no retraining or extra tool calls) and is
the primary driver of model's performance gains on hard examples. 

\subsection{Training pipeline}
\label{sec:training}

\noindent \textbf{Base model.}
We selected Qwen3-Coder-30B-A3B~\cite{qwen3coder} after
evaluating models from 1.1B to 30B parameters. For inference, we serve via vLLM with tensor parallelism and native function-calling.

\noindent \textbf{Stage 1: SFT.}
We fine-tune Qwen3-Coder-30B-A3B on 647
curated episodes using LoRA~\cite{hu2022lora} (rank 32, alpha
64, 26.7M params / 0.09\%) for 2 epochs with
lr=$2{\times}10^{-4}$, batch=4, and BF16 on 8$\times$A100 40GB.

\noindent \textbf{Stage 2: GRPO.}
From the SFT checkpoint, we apply GRPO~\cite{shao2024deepseek}.
For each prompt, $G{=}4$
kernels are generated, each scored with binary reward ($+1.0$
correct, $-0.5$ fail), and group-relative advantages computed.
We trained for 200 steps with lr=$5{\times}10^{-7}$ and $\beta{=}0.1$. 

\section{Experiments}
\label{sec:experiments}

\begin{table}[t]
\centering
\caption{\textbf{Main results on \nkibench{} (150-task).} Pass
rate on balanced subset (50 per level). All results from real
Trn1 hardware. Note: at $N{=}150$, differences of 1--2 tasks
($\sim$0.7pp) are within noise; per-level and 60-task results
(Tables~\ref{tab:by_level}, \ref{tab:single_multi}) show clearer
separation.}
\label{tab:main_results}
\scriptsize
\setlength{\tabcolsep}{4pt}
\begin{tabular}{l cc}
\toprule
\textbf{Model} & \textbf{Single-Shot} & \textbf{NKIAgent (10 turns)} \\
\midrule
Base (Qwen3-30B-A3B) & 14.0\% (21/150) & 20.0\% (30/150) \\
SFT & 19.3\% (29/150) & \textbf{20.7\% (31/150)} \\
GRPO & 15.3\% (23/150) & 17.3\% (26/150) \\
\bottomrule
\end{tabular}
\end{table}

\begin{table*}[t]
\centering
\caption{\textbf{Pass rate by difficulty level} (150-task,
50 per level). NKIAgent tool use uniquely enables Level~2 and
Level~3 tasks where single-shot achieves near-zero success.}
\label{tab:by_level}
\small
\begin{tabular}{l ccc ccc}
\toprule
& \multicolumn{3}{c}{\textbf{Single-Shot}} & \multicolumn{3}{c}{\textbf{NKIAgent}} \\
\cmidrule(lr){2-4} \cmidrule(lr){5-7}
\textbf{Model} & L1 & L2 & L3 & L1 & L2 & L3 \\
\midrule
Base    & 34\% & 8\% & 0\% & 26\% & 18\% & 16\% \\
SFT  & \textbf{50\%} & 8\% & 0\% & 30\% & 16\% & 16\% \\
GRPO    & 44\% & 2\% & 0\% & 22\% & 18\% & 12\% \\
\midrule
Opus 4.8 & 10\% & 8\% & 0\% & \textbf{84\%} & \textbf{74\%} & \textbf{74\%} \\
\bottomrule
\end{tabular}

\smallskip
\end{table*}

\subsection{\nkibench{}}
\label{sec:nkibench}

\nkibench{}\footnote{Not to be confused with the concurrently
developed \textsc{NKIBench} of~\citet{accelopt}, a kernel
\emph{optimization} suite for Trainium; \nkibench{} targets
kernel \emph{generation} and correctness.} contains 250 tasks
across three difficulty levels:
\textbf{Level~1} (100 tasks: single operations such as matmul,
activations, reductions, convolutions), \textbf{Level~2} (100
tasks: fused operations like conv+relu+bias, attention blocks, MLP
components), and \textbf{Level~3} (50 tasks: full model
components including Transformer layers, ResNet bottlenecks, Mamba
blocks). Each task provides a PyTorch \texttt{Model} class and
\nki{}-specific metadata. Of 250 tasks, $\sim$150 are adapted
from KernelBench, 60 require decomposition, and 40 are
\nki{}-native. We evaluate on a balanced 150-task subset (50 per
level) and a 60-task subset for ablations.

\subsection{Evaluation setup}
\label{sec:setup}

\textbf{Models:} (1)~Base Qwen3-Coder-30B-A3B; (2)~SFT; (3)~GRPO (SFT + 200 RL steps); (4)~Claude Sonnet
4; (5)~Claude Opus~4.8 (frontier upper bound). All Claude models
use identical \ours{} tools.

\textbf{Modes:} Single-shot (SS, one generation, no tools),
MT-5 (5 turns without tools; the harness compiles the output and
feeds back raw error text as a user message, but the model
cannot invoke tools itself), and NKIAgent ($\leq$10 turns with
compile/verify tools).

\textbf{Metric:} pass rate $=$ compiles on \texttt{neuronx-cc}
$+$ numerically correct (atol=rtol=$10^{-3}$) vs PyTorch
reference. All on real Trn1 hardware.

\subsection{Main results}
\label{sec:main_results}

Tables~\ref{tab:main_results} and~\ref{tab:by_level} present the
main results. On the aggregate 150-task metric, SFT NKIAgent
(20.7\%) and Base NKIAgent (20.0\%) are within statistical noise
(1 task apart). The clearer signal emerges from per-level
breakdown and 60-task ablations
(Table~\ref{tab:single_multi}). Key findings:

\begin{itemize}[leftmargin=*,itemsep=2pt]
    \item \textbf{SFT + NKIAgent} achieves 20.7\% on
    150-task and 25.0\% on 60-task. The \ours{} framework
    scales to 77.3\% with Opus~4.8
    (Table~\ref{tab:single_multi}), showing the framework
    design is sound and model capability is the primary
    bottleneck.
    \item \textbf{Tool use uniquely enables L2/L3.} No model
    solves any L3 task in single-shot (0/50); NKIAgent achieves
    16\% via iterative compile-fix.
    \item \textbf{SFT improves L1}: 34\%$\rightarrow$50\%,
    reflecting learned \nki{} patterns (tile dims,
    \texttt{nl.load}/\texttt{nl.store}).
    \item \textbf{L1 NKIAgent $<$ L1 SS}: a counterintuitive
    pattern where tool use \emph{hurts} easy tasks. The agent
    sometimes ``fixes'' already-correct code after seeing
    unrelated compiler warnings, or exhausts turns exploring
    alternatives. Tool use is net-positive only for tasks that
    \emph{require} iterative debugging (L2/L3).
    \item \textbf{GRPO degrades performance} vs SFT across
    all settings (Section~\ref{sec:grpo_negative}).
\end{itemize}

\begin{figure*}[t]
    \centering
    \includegraphics[width=\textwidth]{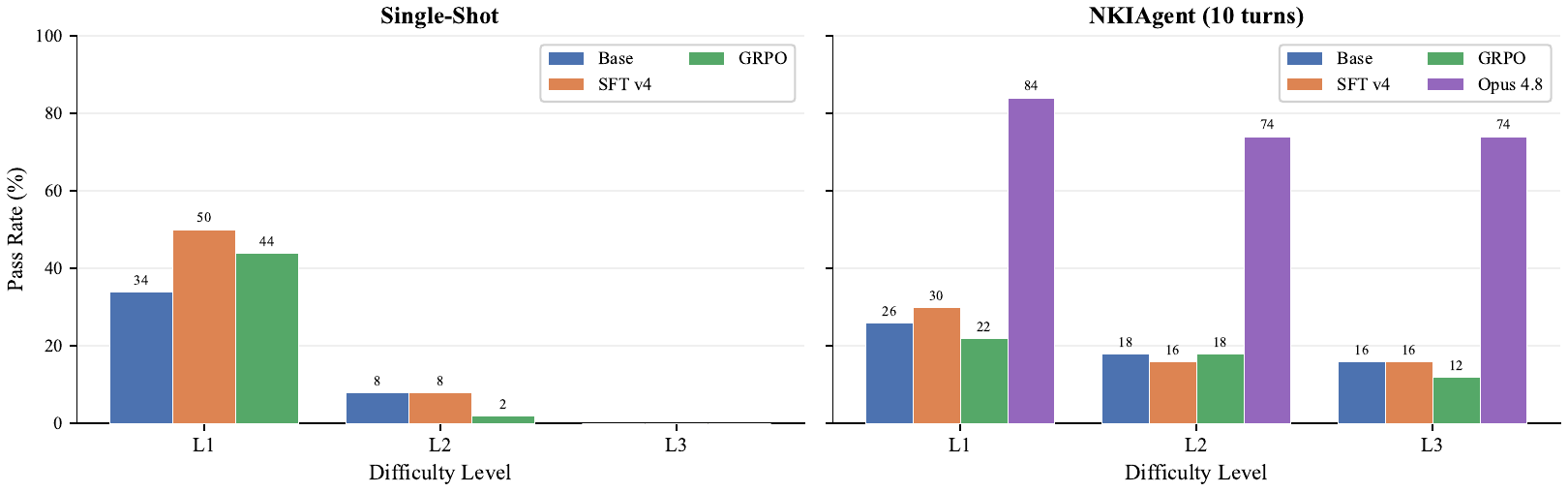}
    \caption{\textbf{Pass rate by difficulty level} (150-task).
    Left: single-shot shows steep degradation from L1 to L3,
    with all trained models at 0\% on L3. Right: NKIAgent enables
    meaningful L2/L3 performance. Opus~4.8 with the rank-aware
    prompt (purple) dominates all levels, particularly L3 (74\%
    vs $\leq$16\% for trained models).}
    \label{fig:level_comparison}
\end{figure*}

\begin{figure*}[t]
    \centering
    \includegraphics[width=\textwidth]{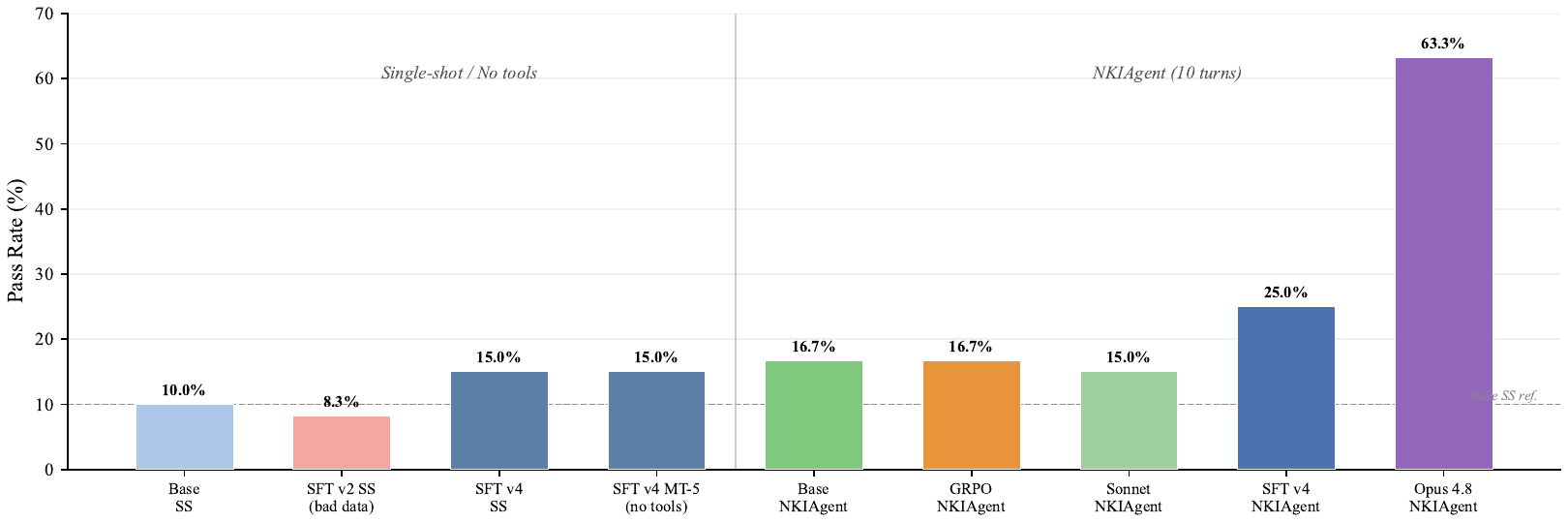}
    \caption{\textbf{Ablation study (60-task subset).} Opus~4.8
    NKIAgent (63.3\%) establishes the framework's upper bound.
    SFT NKIAgent (25.0\%) captures 40\% of frontier
    performance. SFT v2 with bad data (8.3\%) underperforms the
    base model (10.0\%).}
    \label{fig:ablation}
\end{figure*}

\begin{table}[t]
\centering
\caption{\textbf{Ablation study} comparing evaluation modes and
model configurations. The 60-task subset enables additional
comparisons. ``---'' indicates configurations not evaluated on
the larger set due to compute constraints (each 150-task
NKIAgent run requires $\sim$20 hours on Trn1). \textbf{Bold}:
best trained model and best overall.}
\label{tab:single_multi}
\small
\begin{tabular}{l cc}
\toprule
\textbf{Configuration} & \textbf{60-task} & \textbf{150-task} \\
\midrule
Base SS                & 10.0\% (6/60)  & 14.0\% (21/150) \\
SFT SS              & 15.0\% (9/60)  & 19.3\% (29/150) \\
SFT MT-5 (no tools) & 15.0\% (9/60)  & --- \\
\midrule
Base NKIAgent          & 16.7\% (10/60) & 20.0\% (30/150) \\
\textbf{SFT NKIAgent} & \textbf{25.0\% (15/60)} & \textbf{20.7\% (31/150)} \\
GRPO NKIAgent          & 16.7\% (10/60) & 17.3\% (26/150) \\
\midrule
Sonnet NKIAgent        & 15.0\% (9/60)  & --- \\
\midrule
Opus 4.8 SS            & 0.0\% (0/60)   & 6.0\% (9/150) \\
\textbf{Opus 4.8 NKIAgent} & \textbf{63.3\% (38/60)} & \textbf{77.3\% (116/150)} \\
\bottomrule
\end{tabular}
\end{table}

\subsection{Single-shot vs.\ NKIAgent}
\label{sec:single_vs_multi}

Table~\ref{tab:single_multi} reveals several findings.
Multi-turn without tools does \emph{not} help (MT-5 $=$ SS at
15.0\%), echoing the Reflexion~\cite{shinn2023reflexion} finding
that verbal self-feedback without grounded execution signal
yields limited gains. SFT and NKIAgent are super-additive (+5.0pp + +6.7pp
separately, +15.0pp combined). Most strikingly,
\textbf{Opus~4.8 achieves 0\% in single-shot but 63.3\% with
NKIAgent tools} on the 60-task subset (6\% vs 77.3\% on
150-task), the largest tool-use gap in our study,
demonstrating that even frontier models \emph{cannot} generate
correct \nki{} kernels without iterative compiler feedback.
Opus~4.8's L3 pass rate (74\% on 150-task) is particularly
notable: where all other models achieve $\leq$16\%, Opus~4.8
succeeds on nearly three-quarters of full model components.

\paragraph{Cost-capability tradeoff.}
Our SFT model (3B active params, 7.1 tok/s) captures 40\% of
Opus~4.8's performance (25.0\% vs 63.3\% on the 60-task subset)
at $\sim$1/100th the per-token cost. The
SFT model outperforms Sonnet (25.0\% vs 15.0\%) despite
$\sim$300$\times$ less \nki{} training data than CUDA,
demonstrating that domain SFT on scarce data can close a large
fraction of the gap to frontier models.

\section{Analysis}
\label{sec:analysis}

\subsection{SFT data quality progression}
\label{sec:sft_data_quality}

\begin{table*}[t]
\centering
\caption{\textbf{SFT data quality progression.} Each version
fixes a systematic data issue. ``--'' $=$ version discarded
before full evaluation (issues found during spot checks).}
\label{tab:sft_versions}
\small
\begin{tabular}{l l cc}
\toprule
\textbf{Version} & \textbf{Issue Fixed} & \textbf{SS (60)} & \textbf{$\Delta$} \\
\midrule
v1 & Wrong message format                       & --- & --- \\
v2 & 64\% wrong function names + overfitting    & 8.3\% & baseline \\
v3 & 48\% used 3D shapes (should be 2D)         & --- & --- \\
v4 & All fixed: 2D shapes, correct naming       & 15.0\% & +81\% rel. \\
\midrule
Base (no SFT) & ---                              & 10.0\% & --- \\
\bottomrule
\end{tabular}
\end{table*}

Table~\ref{tab:sft_versions} traces our data curation journey.
SFT v2 with bad data hurts performance (8.3\% vs base 10.0\%).
Each fix targeted one issue: v1$\rightarrow$v2 fixed message
format; v2$\rightarrow$v3 standardized function names (64\%
were wrong); v3$\rightarrow$v4 converted 3D tensors to 2D.
This shows that each data quality fix produced larger gains
than any hyperparameter change. 

\subsection{GRPO negative result}
\label{sec:grpo_negative}

\begin{table*}[t]
\centering
\caption{\textbf{GRPO fails to improve over SFT v4.} 200 steps
of GRPO with binary reward degrades performance across all
configurations.}
\label{tab:grpo_negative}
\small
\setlength{\tabcolsep}{8pt}
\begin{tabular}{l cccc}
\toprule
\textbf{Model} & \textbf{SS (60)} & \textbf{NKIAgent (60)} & \textbf{SS (150)} & \textbf{NKIAgent (150)} \\
\midrule
SFT v4 & 15.0\% & \textbf{25.0\%} & \textbf{19.3\%} & \textbf{20.7\%} \\
GRPO   & 13.3\% & 16.7\%          & 15.3\%          & 17.3\% \\
\midrule
$\Delta$ & $-$11\% & $-$33\% & $-$21\% & $-$16\% \\
\bottomrule
\end{tabular}
\end{table*}

\begin{figure*}[!t]
\small
\hrule
\smallskip
\textbf{Turn 1:} Agent generates initial kernel with
\texttt{nl.matmul(a\_tile, b\_tile)} \\
\hspace*{1em}$\hookrightarrow$ \texttt{compile\_kernel}:
\textcolor{red}{FAIL} --- \texttt{nl.matmul} does not exist;
use \texttt{nisa.nc\_matmul} \\
\textbf{Turn 2:} Agent switches to \texttt{nisa.nc\_matmul},
incorrect tile shape (128, 512) \\
\hspace*{1em}$\hookrightarrow$ \texttt{compile\_kernel}:
\textcolor{red}{FAIL} --- free dim of lhs must equal partition
dim of rhs \\
\textbf{Turn 3:} Agent transposes RHS, adds tiling loop over K
dimension \\
\hspace*{1em}$\hookrightarrow$ \texttt{compile\_kernel}:
\textcolor{teal}{OK}. \texttt{verify\_kernel}:
\textcolor{red}{FAIL} --- allclose atol=1e-3 failed \\
\textbf{Turn 4:} Agent adds PSUM accumulation with
\texttt{nisa.nc\_matmul(..., mask=...)} \\
\hspace*{1em}$\hookrightarrow$ \texttt{compile\_kernel}:
\textcolor{teal}{OK}. \texttt{verify\_kernel}:
\textcolor{teal}{PASS} \\
\hrule
\caption{\textbf{Illustrative NKIAgent trace} (condensed from
evaluation logs). The agent iteratively discovers the correct
API (\texttt{nisa.nc\_matmul}), fixes tile dimension
constraints, and adds proper accumulation, errors that are
impossible to recover from in single-shot mode.}
\label{fig:trace}
\end{figure*}

Table~\ref{tab:grpo_negative} shows GRPO consistently
underperforms SFT v4 (average reward flat at $\sim$10--15\%
compile rate across 200 steps). We hypothesize: (1)~binary
reward is too sparse, with no signal between ``almost compiles''
and ``completely wrong''; (2)~subtle code changes flip reward
discontinuously; (3)~when $\sim$75\% of generations fail,
group-relative ranking selects among incorrect outputs.
CUDA-Agent's graded reward (speedup measurements) provides
richer signal. 

\subsection{Error analysis and qualitative findings}
\label{sec:error_analysis}

\paragraph{Category-level patterns.}
SFT excels at elementwise (87.5\%) and activation (90.0\%)
operations. Convolutions, attention, and all Level~3 components
achieve 0\% in single-shot for our model. However, Opus~4.8
NKIAgent achieves 74\% on Level~3 (37/50, 150-task),
demonstrating that the \ours{} tool framework is sufficient for
complex tasks when paired with stronger base models.

\paragraph{Common failures.}
Incorrect tile dimensions (partition $\neq$ 128), unsupported
operation combinations, and SBUF/PSUM memory violations.
Failures are \emph{systematic}: entire categories fail
uniformly, suggesting broader data coverage is more promising
than RL. SFT reliably generates the
load$\rightarrow$compute$\rightarrow$store pattern with
128-element partition but struggles with Tensor Engine matmul
and multi-engine pipelining.

\subsection{Example agent trace}
\label{sec:trace}

Figure~\ref{fig:trace} shows an illustrative NKIAgent
interaction on a matrix multiplication task, condensed from
typical evaluation logs. This trace illustrates why tool use is
essential: the agent must discover that \texttt{nl.matmul} does
not exist in \nki{}, learn the tile shape constraints of
\texttt{nisa.nc\_matmul}, and fix numerical
accumulation, each requiring real compiler feedback.

\section{Limitations and future work}
\label{sec:limitations}

Our work has several limitations. First, we evaluate
correctness only, not runtime performance;
AccelOpt~\cite{accelopt} addresses this complementary axis.
Second, binary reward was insufficient for GRPO; graded rewards
remain future work. Third, all results are single-run without
error bars; small absolute differences may not be statistically
significant. Fourth, our SFT model uses only 3B active
parameters; our Opus~4.8 results (77.3\%) confirm that scaling
model capability significantly improves results within the same
framework. Finally, results are specific to Neuron SDK 2.24.

\section*{Conclusion}
We presented \ours{}, an agentic system combining domain-specific supervised fine-tuning with agentic tool use for kernel generation on AWS Neuron hardware. Evaluated on \nkibench{}, our framework demonstrates that iterative compile-verify-fix loops are essential where even Claude Opus 4.8 scores 6\% in single-shot but 77.3\% with agent tools. Our SFT-trained Qwen3-Coder-30B-A3B captures 40\% of frontier performance at 1/100th the cost, outperforming Claude Sonnet 4. We additionally show that data quality dominates hyperparameter tuning, and that GRPO with binary reward fails to improve over SFT, providing guidance on reward design for RL-based kernel generation.

\balance
\bibliographystyle{icml2026}
\bibliography{references}

\end{document}